\documentclass[10pt,twocolumn,letterpaper]{article}

\usepackage{cvpr}
\usepackage{times}
\usepackage{epsfig}
\usepackage{graphicx}
\usepackage{amsmath,lipsum}
\usepackage{amssymb}
\usepackage{multirow}
\usepackage{color}
\usepackage{slashbox}
\usepackage{arydshln}
\def\infinity{\rotatebox{90}{8}}

\newcommand{\mypm}{\mathbin{\smash{%
\raisebox{0.35ex}{%
            $\underset{\raisebox{0.5ex}{$\smash -$}}{\smash+}$%
            }%
        }%
    }%
}

\newcommand\blfootnote[1]{%
  \begingroup
  \renewcommand\thefootnote{}\footnote{#1}%
  \addtocounter{footnote}{-1}%
  \endgroup
}

\usepackage[pagebackref=true,breaklinks=true,letterpaper=true,colorlinks,bookmarks=false]{hyperref}
\cvprfinalcopy 

\def\cvprPaperID{135} 

\ifcvprfinal\fi
\begin{document}
\pagenumbering{gobble}
\title{NAG: Network for Adversary Generation}
\author{Konda Reddy Mopuri*, Utkarsh Ojha*, Utsav Garg and R. Venkatesh Babu\\
Video Analytics Lab, Computational and Data Sciences\\
Indian Institute of Science, Bangalore, India\\
}

\maketitle

\begin{abstract}
   Adversarial perturbations can pose a serious threat for deploying machine learning systems. Recent works have shown existence of image-agnostic perturbations that can fool classifiers over most natural images. Existing methods present optimization approaches that solve for a fooling objective with an imperceptibility constraint to craft the perturbations. However, for a given classifier, they generate one perturbation at a time, which is a single instance from the manifold of adversarial perturbations. Also, in order to build robust models, it is essential to explore the manifold of adversarial perturbations. In this paper, we propose for the first time, a generative approach to model the distribution of adversarial perturbations. The architecture of the proposed model is inspired from that of GANs and is trained using fooling and diversity objectives. Our trained generator network attempts to capture the distribution of adversarial perturbations for a given classifier and readily generates a wide variety of such perturbations. Our experimental evaluation demonstrates that perturbations crafted by our model (i) achieve state-of-the-art fooling rates, (ii) exhibit wide variety and (iii) deliver excellent cross model generalizability. Our work can be deemed as an important step in the process of inferring about the complex manifolds of adversarial perturbations.
\end{abstract}

\section{Introduction}
\label{sec:intro}
Machine learning\blfootnote{* Equal contribution.} systems are shown~\cite{prsystemsunderattack-pari-2014,evasion-mlkd-2013,adversarialml-acmmm-2011} to be vulnerable to adversarial noise: small but structured perturbation added to the input that affects the model's prediction drastically. Recently, the most successful Deep Neural Network based object classifiers have also been observed~\cite{intriguing-iclr-2014,explainingharnessing-iclr-2015,deepfool-cvpr-2016,atscale-iclr-2017, practical-asiaccs-2017} to be susceptible to adversarial attacks with almost imperceptible perturbations. Researchers have attempted to explain this intriguing aspect via hypothesizing linear behaviour of the models~(\eg~\cite{explainingharnessing-iclr-2015,deepfool-cvpr-2016}), finite training data~(\eg~\cite{deeparch-FTML-2009}), etc. More importantly, the adversarial perturbations exhibit cross model generalizability. That is, the perturbations learned on one model can fool another model even if the second model has a different architecture or has been trained with different dataset~\cite{intriguing-iclr-2014,explainingharnessing-iclr-2015}.

Recent startling findings by Moosavi-Dezfooli \textit{et al.}~\cite{universal-cvpr-2017} and Mopuri \textit{et al.}~\cite{mopuri-bmvc-2017,gduap-arxiv-2018} have shown that it is possible to mislead multiple state-of-the-art deep neural networks over most of the images by adding a single perturbation. That is, these perturbations are image-agnostic and can fool multiple diverse networks trained on a target dataset. Such perturbations are named ``Universal Adversarial Perturbations" (UAP), because a single adversarial noise can perturb images from multiple classes. On one side, the adversarial noise poses a severe threat for deploying machine learning based systems in the real world. Particularly, for the applications that involve safety and privacy (\eg, autonomous driving and access granting), it is essential to develop robust models against adversarial attacks. On the other side, it also poses a challenge to our understanding of these models and the current learning practices. Thus, the adversarial behaviour of the deep learning models to small and structured noise demands a rigorous study now more than ever.
\begin{figure*}[th]
        \centering
        \includegraphics[width=\textwidth]{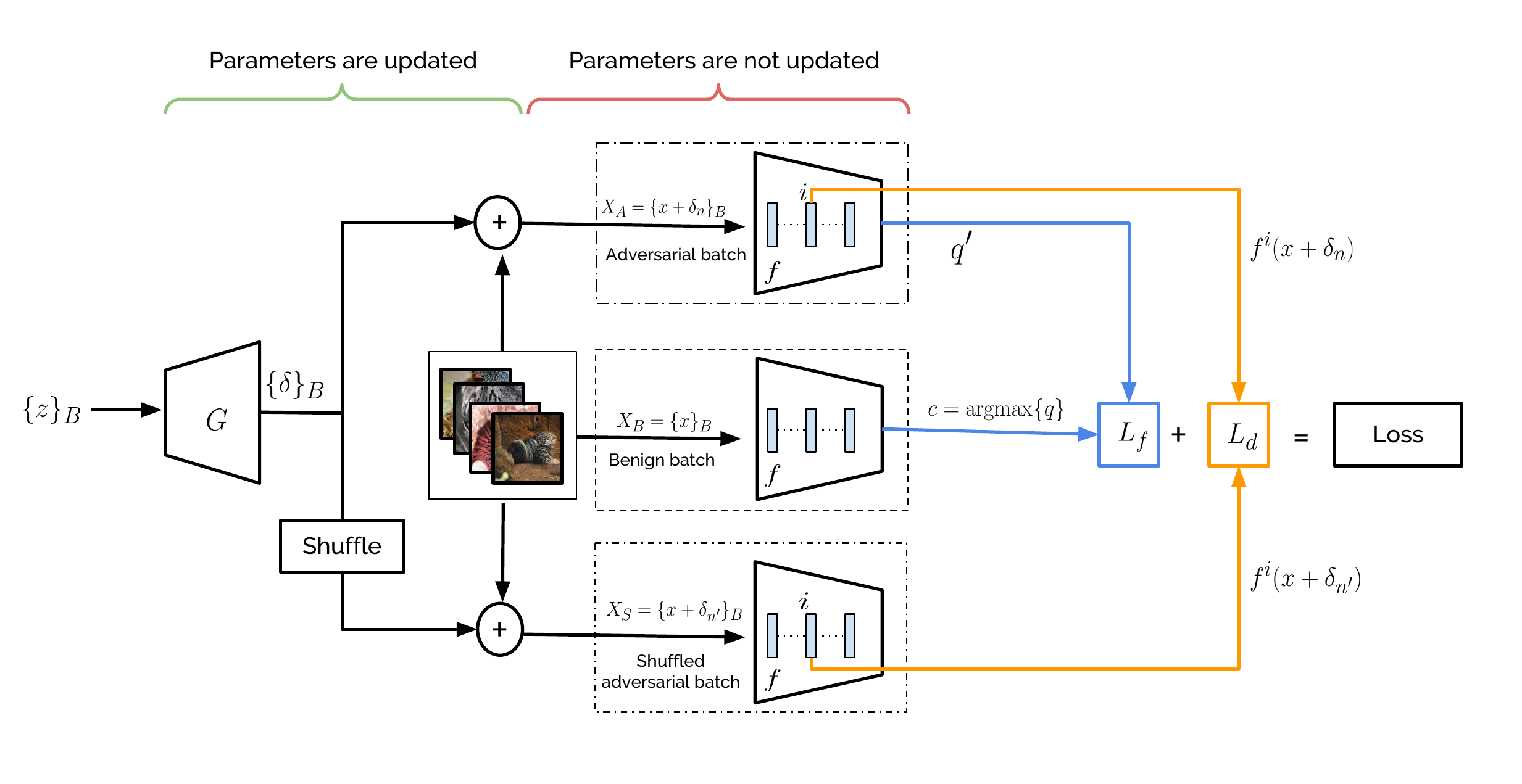}
        \caption{Overview of the proposed approach that models the distribution of universal adversarial perturbations for a given classifier. The illustration shows a batch of $B$ random vectors $\{ z\}_B$ transforming into perturbations $\{\delta\}_B$ by $G$ which get added to the batch of data samples $\{x\}_B$. The top portion shows \textit{adversarial batch} $(X_A)$, bottom portion shows \textit{shuffled adversarial batch} $(X_S)$ and middle portion shows the \textit{benign batch} $(X_B)$. The \textit{Fooling objective} $L_f$ (eq.~\ref{eqn:l_f}) and \textit{Diversity objective} $L_d$ (eq.~\ref{eqn:l_d}) constitute the loss. Note that the target CNN $(f)$ is a trained classifier and its parameters are not updated during the proposed training. On the other hand, the parameters of generator $(G)$ are randomly initialized and learned through backpropagating the loss. (Best viewed in color).
        }
        \label{fig:overview}        
\end{figure*}

All the existing methods, weather image specific~\cite{intriguing-iclr-2014,explainingharnessing-iclr-2015,deepfool-cvpr-2016,carlini-ssp-2017,liu-iclr-2017} or agnostic~\cite{universal-cvpr-2017,mopuri-bmvc-2017,gduap-arxiv-2018}, can craft only a single perturbation that makes the target classifier susceptible. 
Specifically, these methods typically learn a single perturbation $(\delta)$ from a possibly bigger set of perturbations $(\Delta)$ that can fool the target classifier. It is observed that for a given technique (\eg, UAP~\cite{universal-cvpr-2017}, FFF~\cite{mopuri-bmvc-2017}), the perturbations learned across multiple runs are not very different. In spite of optimizing with a different data ordering or initialization, their objectives end up learning very close perturbations in the space (refer sec.~\ref{subsec:diversity}). In essence, these approaches can only prove that the UAPs exist for a given classifier by crafting one such perturbation $(\delta)$. This is very limited information about the underlying distribution of such perturbations and in turn about the target classifier itself. Therefore, a more relevant task at hand is to model the distribution of adversarial perturbations. Doing so can help us better analyze the susceptibility of the models against adversarial perturbations. Furthermore, modelling the distributions would provide insights regarding the transferability of adversarial examples and help to prevent black-box attacks~\cite{distillation-sp-2016,liu-iclr-2017}. It also helps to efficiently generate large number of adversarial examples for learning robust models via adversarial training~\cite{ensembleAT-iclr-2018}.

Empirical evidence~\cite{explainingharnessing-iclr-2015,warde2016} has shown that the perturbations exist in large contiguous regions rather than being scattered in multiple small discontinuous pockets. In this paper, we attempt to model such regions for a given classifier via generative modelling. We introduce a GAN~\cite{gan-nips-2014} like generative model to capture the distribution of the unknown adversarial perturbations. The freedom from parametric assumptions on the distribution and the target distribution being unknown (no known samples from the target distribution of adversarial perturbations) make the GAN framework a suitable choice for our task.

The major contributions of this work are:
\begin{itemize}
    \item A novel objective~(eq. \ref{eqn:l_f}) to craft universal adversarial perturbations for a given classifier that achieves the state-of-the art fooling performance on multiple CNN architectures trained for object recognition.
    \item For the first time, we show that it is possible to model the distribution ($\Delta$) of such perturbations for a given classifier via a generative model. For this, we present an easily trainable framework for modelling the unknown distribution of perturbations.
    \item We demonstrate empirically that the learned model can capture the distribution of perturbations and generates perturbations that exhibit diversity, high fooling capacity and excellent cross model generalizability.
\end{itemize}

The rest of the paper is organized as follows: section~\ref{sec:proposed} details the proposed method, section~\ref{sec:expts} presents comprehensive experimentation to validate the utility of the proposed method, section~\ref{sec:relworks} briefly discusses the existing related works, and finally section~\ref{sec:conclu} concludes the paper.
%
%
\section{Proposed Approach}
\label{sec:proposed}
This section presents a detailed account of the proposed method. For ease of reference, we first briefly introduce the GAN~\cite{gan-nips-2014} framework.
\subsection{GANs}
\label{subsec:gan}
Generative models for images have seen renaissance lately, especially because of the availability of large datasets~\cite{imagenet-ijcv-2015,places-pami-2017} and the emergence of deep neural networks. Particularly, Generative Adversarial Networks (GAN)~\cite{gan-nips-2014} and Variational Auto-Encoders (VAE)~\cite{vae-iclr-2013} have shown significant promise in this direction. In this work, we utilize a GAN like framework to model the distribution of the adversarial perturbations. 

A typical GAN framework consists of two parts: a Generator $(G)$ and a Discriminator $(D)$. The generator $G$ transforms a random vector $z$ into a meaningful image $I$; i.e., $G(z)=I$, where $z$ is usually sampled from a simple distribution (\eg, $\mathcal{N}(0,1)$, $\mathcal{U}(-1,1)$). $G$ is trained to produce images $(I)$ that are indistinguishable from real images from the true data distribution $p_{data}$. The discriminator $D$ accepts an image and outputs the probability for it to be a real image, a sample from $p_{data}$. Typically, $D$ is trained to output low probability $p_D$ when a fake (generated) image is presented. Both $G$ and $D$ are trained adversarially to compete with and improve each other. A properly trained generator $G$ at the end of training is expected to produce images that are indistinguishable from real images.
\subsection{Modelling the adversaries}
\label{subsec:nag}
A broad overview of our method is illustrated in Fig.~\ref{fig:overview}. We first formalize the notations used in the subsequent sections of the paper. Note that in this paper, we have considered CNNs that are trained for object recognition~\cite{resnet-cvpr-2016,googlenet-cvpr-2015,vgg-arxiv-2014,caffe-acmmm-2014}. The data distribution over which the classifiers are trained is denoted as $\mathcal{X}$ and a particular sample from $\mathcal{X}$ is represented as $x$. The target CNN is denoted as $f$, therefore the output of a given layer $i$ is denoted as $f^i(x)$. The predicted label for a given data sample $x$ is denoted as $\hat {k}(x)$. Output of the softmax layer is denoted as $q$, which is a vector of predicted probabilities $q_j$ for each of the target categories $j$. The image-agnostic additive perturbation that fools the target CNN is denoted as $\delta$. $\xi$ denotes the limit on the perturbation $(\delta)$ in terms of its $l_{\infinity}$ norm. Our objective is to model the distribution of such perturbations $(\Delta)$ for a given classifier. Formally, we seek to model
\begin{equation}
\begin{split}
\Delta = \{\delta :  \hat {k}(x+\delta)\ne \hat {k}(x) \text{ for }  x \thicksim \mathcal{X}  \text{ and } \\ ||\: \delta\: ||_{\infinity} < \xi \}
\end{split}
\label{eqn:objective}
\end{equation}
Since our objective is to model the unknown distribution of image-agnostic perturbations for a given trained classifier (target CNN), we make suitable changes in the GAN framework. The modifications we make are: (i) Discriminator $(D)$ is replaced by the target CNN $(f)$ which is already trained and whose weights are frozen, and (ii) a novel loss (fooling and diversity objectives) instead of the adversarial loss to train the Generator $(G)$. Thus, the objective of our work is to train a model $(G)$ that can fool the target CNN. 
The architecture for $G$ is also similar to that of typical GAN which transforms a random sample to an image through a dense layer and a series of deconv layers. More details about the exact architecture are discussed in section~\ref{sec:expts}. We now proceed to discuss the fooling objective that enables us to model the adversarial perturbations for a given classifier.
\subsection{Fooling Objective}
\label{subsec:foolingobjective}
In order to fool the target CNN, the generator $G$ should be driven by a suitable objective. Typical GANs use adversarial loss to train their $G$. However, in this work we attempt to model a distribution whose samples are unavailable. We know only a single attribute of those samples which is \textit{to be able to fool the target classifier}. We incorporate this attribute via a fooling objective to train our $G$ that models the unknown distribution $(\Delta)$ of perturbations.

We denote the label predicted by the target CNN on a clean sample $x$ as \textit{benign prediction} $(c)$ and that predicted on the corresponding perturbed sample $(x+\delta)$ as \textit{adversarial prediction}. Similarly, we denote the output vector of the softmax layer without $\delta$ and after adding $\delta$ as $q$ and $q'$ respectively. Ideally a perturbation $\delta$ should confuse the classifier so as to flip the benign prediction into a different adversarial prediction. For this to happen, after adding $\delta$, the confidence of the benign prediction $(q'_c)$ should be reduced and that of another category should be made higher. Thus, we formulate a fooling loss to minimize the confidence of benign prediction on the perturbed sample $(x+\delta)$
\begin{equation}
L_f = -log(1-q'_c)
\label{eqn:l_f}
\end{equation}
Fig.~\ref{fig:overview} shows a graphical explanation of the objective, where the fooling objective is shown by the blue colored block. Note that the fooling loss essentially encourages $G$ to generate perturbations that decrease confidence of benign predictions and thus eventually flip the label.
\subsection{Diversity Objective}
\label{subsec:diversityloss}
The fooling loss only encourages to learn a $G$ that can guarantee high fooling capability for the generated perturbations $(\delta)$. This objective might lead to some local minima where the $G$ learns only a limited set of effective perturbations as in~\cite{universal-cvpr-2017,mopuri-bmvc-2017}. However, our objective is to model the distribution $\Delta$ such that it covers all varieties of those perturbations. Therefore, we introduce an additional component to the loss that encourages $G$ to explore the space of perturbations and generate a diverse set of perturbations. We term this objective the \textit{Diversity objective}. Within a mini-batch of generated perturbations, this objective indirectly encourages them to be different by separating their feature embeddings projected by the target classifier. In other words, for a given pair of generated perturbations $\delta_n$ and $\delta_n'$, our objective increases the distance between $f^i(x+\delta_n)$ and $f^i(x+\delta_n')$ at a given layer $i$ in the classifier.
\begin{equation}
L_d = -\sum_{n=1 }^B  d( f^i(x_n+\delta_n) , f^i(x_n+\delta_{n'}) )
\label{eqn:l_d}
\end{equation}
where  $n' \text{ is a random index in } [1,B] \text{ and }n' \neq n$,
$B$ is the batch size, $x_n$, $\delta_n$ are $n^{th}$ data sample and perturbation in the mini-batch respectively. Note that a batch contains $B$ perturbations $(\delta)$ generated by $G$ (via transforming random vectors $(z)$) and $B$ data samples $(x)$. $f^{i}$ is the output of the CNN at $i^{th}$ layer and $d(.,.)$ is a distance metric (\eg, Euclidean) between a pair of features. The orange colored block in Fig.~\ref{fig:overview} illustrates the diversity objective.

Thus, our final loss becomes the summation of both fooling and diversity objectives and is given by
\begin{equation}
Loss = L_f + \lambda L_d
\label{eqn:loss}
\end{equation}
Since it is important to learn diverse perturbations that exhibit maximum fooling, we give equal importance to both $L_f$ and $L_d$ in the final loss to learn the $G$ (i.e., $\lambda=1$).
\subsection{Architecture and Implementation details}
\label{subsec:arch-details}
\begin{table*}[]
\footnotesize
\centering
\caption{Average fooling rates of the perturbations modelled by our generative network vs. UAP~\cite{universal-cvpr-2017}. Rows indicate the target net for which perturbations are modelled and columns indicate the net under attack. Note that, in each row, entry where the target CNN matches with the network under attack represents white-box attack and the rest represent the black-box attacks. For our method, along with average fooling rates, the corresponding standard deviations are also mentioned. The best result for each case is shown in bold and UAP best cases are shown in \textit{blue}. Mean avg. fooling rate achieved by the Generator $(G)$ for each of the target CNNs is shown in the rightmost column.}
\label{tab:fooling}
\begin{tabular}{|l|l|l|l|l|l|l|l|l|l|}
\cline{3-10}
\multicolumn{2}{l|}{}       & VGG-F          & CaffeNet       & GoogLeNet      & VGG-16         & VGG-19         & ResNet-50      & ResNet-152    & Mean FR \\ \cline{3-10} \hline 
\multirow{2}{*}{VGG-F}      & Our & \textbf{94.10} $\mypm$ 1.84 & \textbf{81.28}$\mypm$ 3.50          & \textbf{64.15}$\mypm$3.41          & \textbf{52.93}$\mypm$8.50          & \textbf{55.39}$\mypm$2.68          & 50.56$\mypm$4.50          & \textbf{47.67}$\mypm$4.12 &  \textbf{63.73}         \\ \cline{2-10} 
                            & UAP & 93.7           & 71.8           & 48.4           & 42.1           & 42.1           &         -       & 47.4          & 57.58\\ \hline \hline
\multirow{2}{*}{CaffeNet}   & Our & \textbf{79.25}$\mypm$1.44          & \textbf{96.44}$\mypm$1.56 & \textbf{66.66}$\mypm$1.84          & \textbf{50.40}$\mypm$5.61          & \textbf{55.13}$\mypm$4.15          & 52.38$\mypm$3.96          & \textbf{48.58}$\mypm$4.25  &    \textbf{64.12}    \\ \cline{2-10} 
                            & UAP & 74.0           & 93.3           & 47.7           & 39.9           & 39.9           &         -       & 48.0 &  56.71         \\ \hline \hline
\multirow{2}{*}{GoogLeNet}  & Our & \textbf{64.83}$\mypm$0.86          & \textbf{70.46}$\mypm$2.12          & \textbf{90.37}$\mypm$1.55 & \textbf{56.40}$\mypm$4.13          & \textbf{59.14}$\mypm$3.17          & 63.21$\mypm$4.40          & \textbf{59.22}$\mypm$1.64 &    \textbf{66.23}     \\ \cline{2-10} 
                            & UAP & 46.2           & 43.8           & 78.9           & 39.2           & 39.8           &         -       & 45.5  &    48.9      \\ \hline \hline
\multirow{2}{*}{VGG-16}     & Our & 60.56$\mypm$2.24          & \textbf{65.55}$\mypm$6.95          & \textbf{67.38}$\mypm$4.84          & 77.57$\mypm$2.77 & \textbf{73.25}$\mypm$1.63          & 61.28$\mypm$3.47          & 54.38$\mypm$2.63        & \textbf{65.71} \\ \cline{2-10} 
                            & UAP & {\color{blue}\textbf{63.4}}           & 55.8           & 56.5           & {\color{blue}\textbf{78.3}}           & 73.1           &       -         & {\color{blue}\textbf{63.4}}    &   65.08    \\ \hline \hline
\multirow{2}{*}{VGG-19}     & Our & \textbf{67.80}$\mypm$2.49          & \textbf{67.58}$\mypm$5.59          & \textbf{74.48}$\mypm$0.94          & \textbf{80.56}$\mypm$3.26          & \textbf{83.78}$\mypm$2.45 & 68.75$\mypm$3.38          & \textbf{65.43}$\mypm$1.90      &   \textbf{72.62} \\ \cline{2-10} 
                            & UAP & 64.0           & 57.2           & 53.6           & 73.5           & 77.8           &          -      & 58.0 &   64.01       \\ \hline \hline
\multirow{2}{*}{ResNet-50}  & Our & 47.06$\mypm$2.60          & 63.35$\mypm$1.70          & 65.30$\mypm$1.14          & 55.16$\mypm$2.61          & 52.67$\mypm$2.58          & 86.64$\mypm$2.73 & 66.40$\mypm$1.89 &   62.37      \\ \cline{2-10} 
                            & UAP &  -              &    -            &      -          &        -        &      -          &            -    &      - &    -     \\ \hline \hline
\multirow{2}{*}{ResNet-152} & Our & \textbf{57.66}$\mypm$4.37          & \textbf{64.86}$\mypm$2.95         & \textbf{62.33}$\mypm$1.39          & \textbf{52.17}$\mypm$3.41          & \textbf{53.18}$\mypm$4.16          & 73.32$\mypm$2.75          & \textbf{87.24}$\mypm$2.72 & \textbf{64.39}\\ \cline{2-10} 
                            & UAP & 46.3           & 46.3           & 50.5           & 47.0           & 45.5           &  -              & 84.0 &     53.27      \\ \hline
\end{tabular}
\end{table*}
Before we present the experimental details, we describe the implementation and working details of the proposed architecture. The generator part $(G)$ of the network maps the latent space $Z$ to the distribution of perturbations $(\Delta)$ for a given target classifier. The architecture of the generator consists of $5$ deconv layers. The final deconv layer is followed by a $tanh$ non-linearity and scaling by $\xi$. Doing so restricts the perturbations' range to $\bigl[-\xi,\: \xi\bigr]$. Following~\cite{universal-cvpr-2017,mopuri-bmvc-2017,gduap-arxiv-2018}, the value of $\xi$ is chosen to be $10$ in order to add a quasi-imperceptible adversarial noise. The generator network is adapted from ~\cite{improvedgans-nips-2016}. We performed all our experiments on a variety of CNN architectures trained to perform object recognition task on the ILSVRC-$2014$~\cite{imagenet-ijcv-2015} dataset. We kept the architecture of our generator $(G)$ unchanged for different target CNN architectures and separately learned the corresponding adversarial distributions.

During training, we sample a batch of random vectors $z \in \mathcal{R}^d$ 
from the uniform distribution $\mathcal{U} \bigl[ -1, 1 \bigr]$ which in turn get transformed by $G$ into a batch of perturbations $\{\delta\}_B=\{\delta_1, \delta_2, \delta_3,\ldots, \delta_B\}$ each of size equal to that of the image (\eg, $224 \times 224 \times 3$). We also sample $B$ images $\{x\}_B=\{x_1, x_2, x_3, \ldots, x_B \} $ from the available training data to form the mini-batch training data, denoted as \textit{benign batch }$(X_B)$. We now add the perturbations to the training data in a one-to-one manner i.e. one perturbation gets added to the corresponding image of the batch, which gives us the \textit{adversarial batch}, $X_A=\{x_1+\delta_1, x_2+\delta_2, x_3+\delta_3,\ldots,x_B+\delta_B\}$. This is shown in the top portion of Fig.~\ref{fig:overview}. We also randomly shuffle the perturbations ensuring no perturbation remains in its original index in the batch, i.e., $\{ {\delta_1}{'}, {\delta_2}', {\delta_3}', \ldots, {\delta_B}'\}$ such that $\delta_i\neq {\delta_i}',\: \forall i$. With this, we form a \textit{shuffled adversarial batch} as $X_S=\{x_1+{\delta_1}', x_2+{\delta_2}', x_3+{\delta_3}',\ldots,x_B+{\delta_B}'\}$, which is shown in the bottom portion of Fig.~\ref{fig:overview}. Note that in order to prepare $X_S$, only the perturbations $(\{\delta\}_B)$ are shuffled but not the data samples $(\{x\}_B)$.

Thus, each of our training iterations consists of three  quasi-batches, namely, (i) Benign images batch $X_B$, (ii) Adversarial batch $X_A$, and (iii) Shuffled adversarial batch $X_S$. These are the three portions shown in Fig.~\ref{fig:overview}. We now feed these through the target CNN $(f)$ and compute the loss. We obtain the benign predictions $(c)$ over the clean batch samples $\{x\}_B$. These labels are used to compute the confidences $(q'_c)$ for the corresponding adversarial batch samples. This forms the fooling objective as shown in eq.~\ref{eqn:l_f}. Similarly, we obtain the feature representations at the softmax layer (probability vectors) for both adversarial and shuffled adversarial batches (top and bottom portions of Fig.~\ref{fig:overview}) to compute the diversity component of the loss as shown in eq.~\ref{eqn:l_d}. Essentially, our diversity objective pushes apart the final layer representations corresponding to two different perturbations ($\delta_i$  and ${\delta_i}'$) via maximizing the cosine distance between them.

Note that we update only the $G$ part of the network and the target CNN, which is a pretrained classifier under attack, remains unchanged. We iteratively perform the loss computation and parameter updation for all the samples in the training data. During training, we use a small held-out set of $1000$ random images as validation set and stop our training upon reaching best performance on this set. In our experiments, the maximum number of epochs is set to $100$ and we observe that training of generators for all the target CNNs gets saturated at around $60-70$ epochs.
\section{Experiments}
\label{sec:expts}
For all our experiments, we worked with $10000$ ($10$ per category, similar to~\cite{universal-cvpr-2017}) training images randomly chosen from ILSVRC $2014$ train set and $50000$ images of ILSVRC $2014$ validation set as our testing images. The latent space dimension $d$ is set to $10$. We have experimented with spaces of different dimensions (\eg, $50$, $100$) and observed that the fooling rates obtained are very close. However, we observe the generated perturbations for $d=10$ demonstrate larger visual variety than other cases. Thus, we keep $d=10$ for all our experiments. We use a batch size $(B)$ of $64$ for shallow networks such as VGG-F~\cite{vggf-bmvc-2014} and GoogLeNet~\cite{googlenet-cvpr-2015}, and $32$ for the rest. The models are implemented in TensorFlow~\cite{tensorflow2015-whitepaper-short} with Adam optimizer~\cite{adam-iclr-2014} on a TITAN-X GPU card. Codes for the project are available at \url{https://github.com/val-iisc/nag}.
\subsection{Perturbations and the fooling rates}
\label{subsec:fooling}
The fooling rates achieved by the perturbations crafted by the learned generative model $(G)$ are presented in Table~\ref{tab:fooling}. Results are shown for seven different network architectures trained on ILSVRC-$2014$ dataset computed for over $50000$ test images. We also investigate the transfer rates of the perturbations by attacking other unknown models along with the target CNN. Rows denote a particular target CNN for which we have modelled the distribution of perturbations and the columns represent the classifiers we attack. Note that in each row, when the target CNN (row) matches with the system under attack (column), the fooling indicates the white-box attack scenario and all other entries represent the black-box attack scenario.

Since our $G$ network models the perturbation space, we can now easily generate a perturbation by sampling a $z$ and feeding it through the $G$. In Table~\ref{tab:fooling}, we report the mean fooling rates after generating multiple perturbations for a given CNN classifier. Particularly, the white-box fooling rates are computed by averaging over $100$ perturbations and black-box rates are averaged over $10$. The standard deviations are mentioned next to the fooling rates. Also, the mean average fooling rate achieved by the learned model $(G)$ for each of the target CNNs is shown in the rightmost column. Clearly, the proposed generative model captures the perturbations with higher fooling rates than UAP~\cite{universal-cvpr-2017}. Note that of all the $36$ entries for which UAP~\cite{universal-cvpr-2017} provided their fooling rates, in only $3$ cases (indicated in bold faced blue in Table~\ref{tab:fooling}) they perform better than us. The mean fooling
rate (of all the entries in the table, except the rightmost column) obtained by the UAP~\cite{universal-cvpr-2017} is $57.66$ and that achieved by our model is $65.68$, which is a significant $8\%$ improvement. 

\begin{figure*}[h]
\centering
\noindent\begin{minipage}{\textwidth}
  \centering
  \begin{minipage}{.13\textwidth}
  	\centering
    \includegraphics[width=\linewidth]{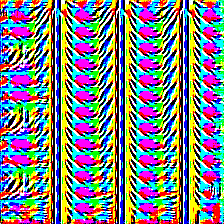}\
    VGG-F
  \end{minipage}
   \begin{minipage}{.13\textwidth}
   	\centering
    \includegraphics[width=\linewidth]{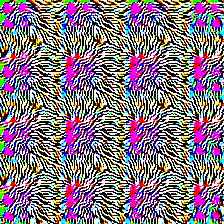}\
    CaffeNet
  \end{minipage}
  \begin{minipage}{.13\textwidth}
  	\centering
    \includegraphics[width=\linewidth]{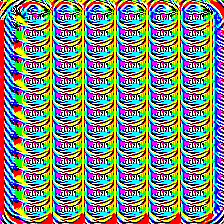}\
    GoogLeNet
  \end{minipage}
  \begin{minipage}{.13\textwidth}
  	\centering
    \includegraphics[width=\linewidth]{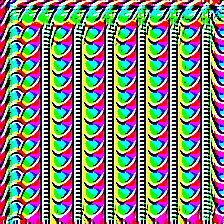}\
    VGG-16
  \end{minipage}
   \begin{minipage}{.13\textwidth}
   	\centering
    \includegraphics[width=\linewidth]{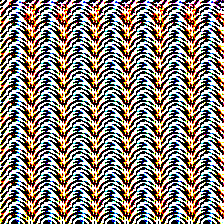}\
    VGG-19
  \end{minipage}
  \begin{minipage}{.13\textwidth}
  	\centering
    \includegraphics[width=\linewidth]{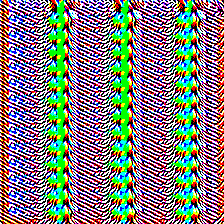}\
     ResNet-50
  \end{minipage}
  \begin{minipage}{.13\textwidth}
  \centering
    \includegraphics[width=\linewidth]{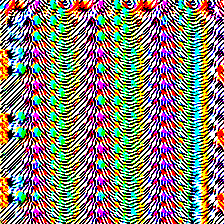}\
    ResNet-152
  \end{minipage}
\end{minipage}
\vspace{0.1cm}
\caption{Sample universal adversarial perturbations for different networks. The target CNN is mentioned below the perturbations. Note that these are one sample from each of the corresponding distributions and across different samplings of the same generative model the perturbations vary visually.}
\label{fig:perturbations}
\end{figure*}

\begin{figure*}[h]
\centering
\noindent\begin{minipage}{\textwidth}
  \centering
  \begin{minipage}{.18\textwidth}
  	\centering
    \includegraphics[width=\linewidth]{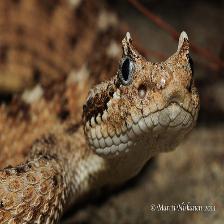}\
    Clean : Horned viper
  \end{minipage}
   \begin{minipage}{.18\textwidth}
   	\centering
    \includegraphics[width=\linewidth]{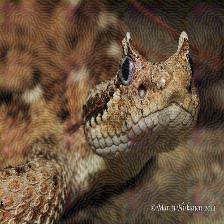}\
    CaffeNet : N. Terrier
    \end{minipage}
  \begin{minipage}{.18\textwidth}
  	\centering
    \includegraphics[width=\linewidth]{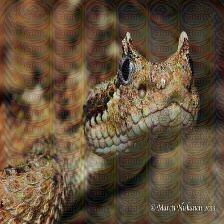}\
    GoogLeNet : Hamper
  \end{minipage}
   \begin{minipage}{.18\textwidth}
   	\centering
    \includegraphics[width=\linewidth]{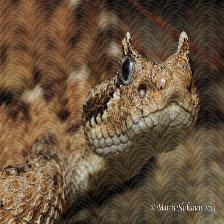}\
    VGG-19 : Hyena
  \end{minipage}
  \begin{minipage}{.18\textwidth}
  	\centering
    \includegraphics[width=\linewidth]{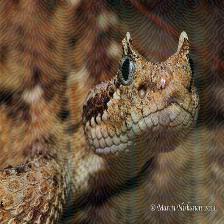}\
     ResNet-152 : Chiton
  \end{minipage}
  \vspace{0.002\textwidth}
\end{minipage}
\vspace{0.1cm}
\caption{A clean image (left most) and corresponding adversarial images crafted for multiple networks along with predictions.}
\label{fig:perturbed-samples}
\end{figure*}

Figure~\ref{fig:perturbations} shows the perturbations generated by the proposed generative model for different target CNNs. Note that each of them is one random sample from the corresponding distributions of perturbations. Fig.~\ref{fig:perturbed-samples} shows a benign sample and the corresponding perturbed samples 
after adding perturbations for multiple CNNs. Note that the perturbations are sampled from the corresponding distributions learned by our method.
\begin{figure*}[ht!]
\centering
\noindent\begin{minipage}{\textwidth}
\begin{minipage}{\textwidth}
  \centering
  \begin{minipage}{.17\textwidth}
  	\centering
    \includegraphics[width=\linewidth]{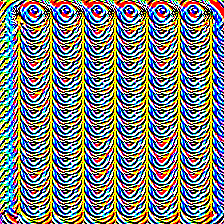}
  \end{minipage}
   \begin{minipage}{.17\textwidth}
   	\centering
    \includegraphics[width=\linewidth]{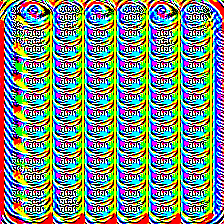}
  \end{minipage}
  \begin{minipage}{.17\textwidth}
  	\centering
    \includegraphics[width=\linewidth]{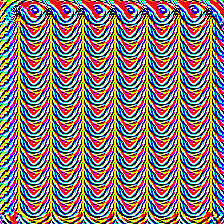}
  \end{minipage}
   \begin{minipage}{.17\textwidth}
   	\centering
    \includegraphics[width=\linewidth]{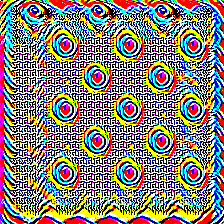}
  \end{minipage}
   \begin{minipage}{.17\textwidth}
   	\centering
    \includegraphics[width=\linewidth]{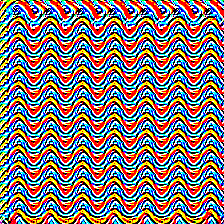}
  \end{minipage}
  \vspace{0.002\textwidth}
\end{minipage}
\vspace{0.002\textwidth}
\end{minipage}
\noindent\begin{minipage}{\textwidth}
  \centering
  \includegraphics[width=0.885\linewidth]{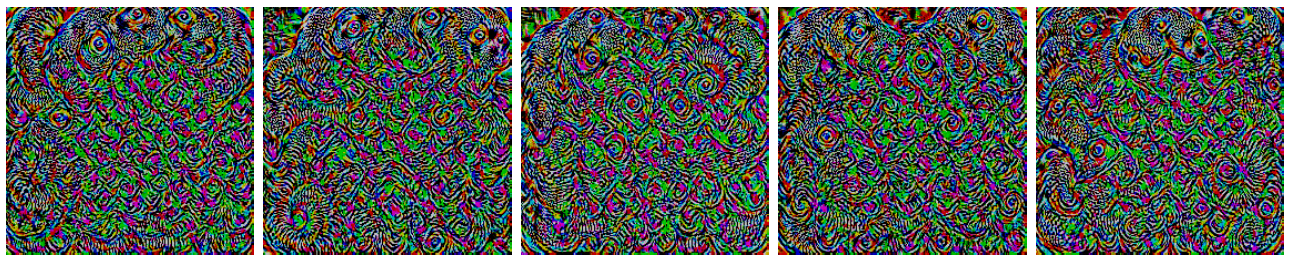}
 \end{minipage}
\vspace{0.1cm}  
\caption{Sample perturbations generated by proposed approach (top) and UAP~\cite{universal-cvpr-2017} (bottom) for GoogLeNet~\cite{googlenet-cvpr-2015}. Note that the perturbations generated by~\cite{universal-cvpr-2017} look very similar to each other, whereas generated by our approach showcase diversity. These results show that the proposed method faithfully models the distribution of perturbations that can effectively fool the target CNN.
}
\label{fig:diversity-googlenet}
\end{figure*}
\begin{figure*}[h]
\centering
\noindent\begin{minipage}{\textwidth}
  \centering
  \begin{minipage}{.18\textwidth}
  	\centering
    \includegraphics[width=\linewidth]{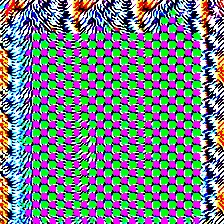}\
    $0.0*z_1+1.0*z_2:87.53$
  \end{minipage}
   \begin{minipage}{.18\textwidth}
   	\centering
    \includegraphics[width=\linewidth]{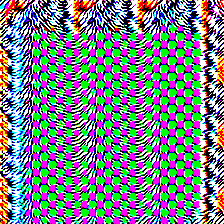}\
    $0.1*z_1+0.9*z_2:87.71$
  \end{minipage}
  \begin{minipage}{.18\textwidth}
  	\centering
    \includegraphics[width=\linewidth]{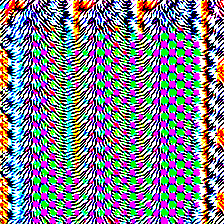}\
    $0.2*z_1+0.8*z_2:87.73$
  \end{minipage}
   \begin{minipage}{.18\textwidth}
   	\centering
    \includegraphics[width=\linewidth]{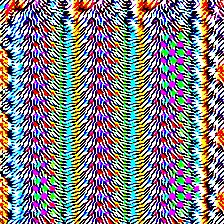}\
    $0.3*z_1+0.7*z_2:87.25$
  \end{minipage}
  \begin{minipage}{.18\textwidth}
  	\centering
    \includegraphics[width=\linewidth]{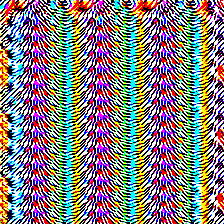}\
     $0.4*z_1+0.6*z_2:87.48$
  \end{minipage}
  \vspace{0.002\textwidth}
\end{minipage}
\noindent\begin{minipage}{\textwidth}
  \centering
  \begin{minipage}{.18\textwidth}
  	\centering
    \includegraphics[width=\linewidth]{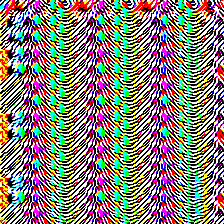}\
    $0.5*z_1+0.5*z_2:88.17$
  \end{minipage}
   \begin{minipage}{.18\textwidth}
   	\centering
    \includegraphics[width=\linewidth]{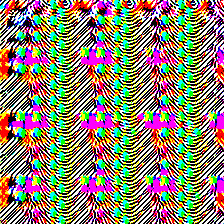}\
    $0.6*z_1+0.4*z_2:87.94$
  \end{minipage}
  \begin{minipage}{.18\textwidth}
  	\centering
    \includegraphics[width=\linewidth]{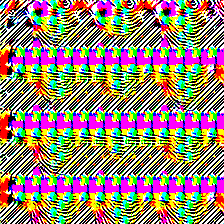}\
    $0.7*z_1+0.3*z_2:87.84$
  \end{minipage}
   \begin{minipage}{.18\textwidth}
   	\centering
    \includegraphics[width=\linewidth]{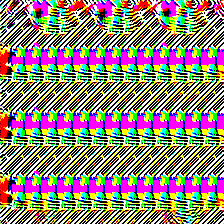}\
    $0.8*z_1+0.2*z_2:88.16$
  \end{minipage}
  \begin{minipage}{.18\textwidth}
  	\centering
    \includegraphics[width=\linewidth]{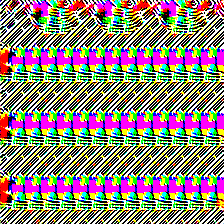}\
     $0.9*z1 + 0.1*z_2:87.34$
  \end{minipage}
  \vspace{0.002\textwidth}
\end{minipage}
\vspace{0.1cm}
\caption{Interpolation between a pair of points in $Z$ space shows that the distribution learned by our generator has smooth transitions. The figure shows the perturbations corresponding to $10$ points on the line joining a pair of points $(z_1 \text{ and } z_2)$ in the latent space. Note that these perturbations are learned to fool the ResNet-152~\cite{resnet-cvpr-2016} architecture. Below each perturbation, the corresponding fooling rate obtained over $50000$ images from ILSVRC $2014$ validation images is mentioned. This shows the fooling capability of these intermediate perturbations is also high and remains same at different locations in the learned distribution of perturbations.}
\label{fig:interpolation}
\vspace{-0.5cm}
\end{figure*}
\subsection{Effect of training data on the modelling}
\label{subsec:effect-training-data}
In this subsection, we examine the effect of available training data on the learning. We have considered ResNet-$152$ model and various training data sizes (equal population from each category). Table~\ref{rebut-tab:effect-training-data} presents the fooling rates obtained by the crafted perturbations in both white-box and black-box setup. Note that the black-box fooling rates are obtained by averaging the fooling rates obtained on three (GoogLeNet, VGG-19 and ResNet-50) CNN models. As one expects, owing to better modelling induced by availability of more training data, the fooling rates increase with available training data.
\begin{table}[]
\centering
\caption{Effect of training data on the modelling.}
\label{rebut-tab:effect-training-data}
\begin{tabular}{|l|l|l|l|l|l|}
\hline
               & $1000$ & $2000$ & $4000$ & $10000$ 
               & $50000$ \\ \hline
White-box      &    61.54  & 73.19     &  78.18    &   87.24    &   91.16    \\ \hline
Mean BBFR &   39.46   &   45.12   &  51.87   &   62.94    &     67.45  \\ \hline
\end{tabular}
\end{table}
\subsection{Diversity of perturbations}
\label{subsec:diversity}
\vspace{-0.2cm}
We examine the diversity of the generated perturbations by our model. 
It can be interesting to examine the predicted label distribution after adding the perturbations. Doing so can reveal if there exist any dominant labels that most of the images get confused to or whether the confusions are diverse. In this subsection, we analyze the labels predicted by the target CNN after adding perturbations modelled by the corresponding generative models $(G)$. We have considered VGG-F architecture and the $50000$ images from the validation set of ILSVRC-$2014$. We compute the mean histogram of the predicted labels for $10$ perturbations generated by our $G$. The top-$5$ categories are: \texttt{\{jigsaw puzzle, maypole, otter, dome, electric fan\}}. Though there exists a slight domination from some of the categories, the extent of domination is far less compared to~\cite{universal-cvpr-2017}. While in our case, $257$ categories account for the $95\%$ of the predicted labels, for UAP~\cite{universal-cvpr-2017}, it is $173$ categories. The $48.6\%$ relative higher diversity compared to~\cite{universal-cvpr-2017} is attributed to the effectiveness of the proposed diversity loss (eq.~\ref{eqn:l_d}) which encourages the model to explore various regions in the adversarial manifold.
\subsection{Traversing the manifold of perturbations}
\label{subsec:interpolation}
In this subsection, we perform experiments to understand the landscape of the latent space $(Z)$. In case of GANs, traversing on the learned manifold generally tells about the signs of memorization~\cite{dcgan-arxiv-2015}. While walking on the latent space, if the image generations result in semantic changes, it is considered that the model has learned relevant and interesting representations. However, our generative model attempts to learn the unknown distribution of adversarial perturbations with no samples from the target distribution. Therefore, it is not relevant to investigate for the smooth semantic changes in the generations but only to look for smooth visual changes, while retaining the ability to fool the target classifier.

Figure~\ref{fig:interpolation} shows the results of interpolation experiments on the ResNet-152~\cite{resnet-cvpr-2016} classifier. We have randomly taken a pair of points $(z_1 \text{ and } z_2)$ in the latent $(Z)$ space and considered $10$ intermediate points on the line joining $z_1$ and $z_2$. We have generated the perturbations corresponding to these intermediate points by feeding them through the learned generative model $(G)$. Figure~\ref{fig:interpolation} shows the generated perturbations at the intermediate points along with their fooling rates. We clearly observe that the perturbations change smoothly between any pair of consecutive points and the sequence gives a morphing like effect with large number of intermediate points.  For each of the intermediate perturbations, fooling rate is computed over the $50000$ images from the ILSVRC-$2014$ validation set. In Fig.~\ref{fig:interpolation}, below each of these perturbations, corresponding fooling rates are mentioned. The high and consistent fooling rates along the path demonstrate that the modelling of the adversarial distribution has been faithful. The proposed approach generates perturbations smoothly from the underlying manifold. We attribute this ability of our learned generative model to the effectiveness of the proposed objectives in the loss.
\subsection{Modelling adversaries for multiple targets}
\label{subsec:multiple-targets}
\begin{table*}[t]
\centering
\caption{Mean fooling rates for $10$ perturbations sampled from the distribution of adversaries modelled for multiple target CNNs. The perturbations result an average fooling rate of $80.02\%$ across the $7$ target CNNs which is higher than the best mean fooling rate of $72.62\%$ achieved by the generator learned for VGG-19.
}
\label{tab:multiple-targets}
\begin{tabular}{|l|l|l|l|l|l|l|l|}
\hline
Network      & VGG-F & CaffeNet & GoogLeNet & VGG-16 & VGG-19 & ResNet-50 & ResNet-152 \\ \hline
Fooling rate & 83.74 & 86.94    & 84.79     & 73.73  & 75.24  & 80.21     & 75.84      \\ \hline
\end{tabular}
\end{table*}
Transferability of the adversarial perturbations (both image specific and agnostic) has been an intriguing revelation by many of the recent adversarial perturbations works~\cite{intriguing-iclr-2014,explainingharnessing-iclr-2015,practical-asiaccs-2017,universal-cvpr-2017,mopuri-bmvc-2017,gduap-arxiv-2018}. Moosavi-Dezfooli \textit{et al.}~\cite{universal-cvpr-2017}  attempted to explain the cross-model generalizability with the correlation among different regions in the classification boundaries learned by them. In this subsection, we investigate if we can learn to model a single distribution of adversaries that fool multiple target CNNs simultaneously. 

We consider all the $7$ target CNNs presented in Table~\ref{tab:fooling} to model a single adversarial manifold that can fool all of them simultaneously. We keep the $G$ part of the proposed architecture unchanged while we replace single target classifier with all the target networks. Because of the memory constraint to fit all the models, we train with a smaller batch size.
The loss to train the generator is the summation of the individual losses (eq.~\ref{eqn:loss}) computed for each of the target CNNs separately. Thus, the objective driving the optimization aims to craft the perturbations that fool all the target CNNs simultaneously. Similar to the single target case, the diversity objective (eq.~\ref{eqn:l_d}) encourages to explore multiple regions covering the manifold of perturbations and model a distribution with a lot of variety.

Table~\ref{tab:multiple-targets} presents the mean fooling rates obtained by $10$ samples from the distribution of perturbations learned to fool all the $7$ target CNNs. The fooling rates are slightly lesser than those obtained for the dedicated optimization (white-box attacks in Tab.~\ref{tab:fooling}). However, given the complexity of the modelling, the learned perturbations achieve a remarkable average fooling rate of $80.07\%$. Note that this is around $8\%$ higher than the best mean fooling rate obtained by an individual network (computed for each row in Tab.~\ref{tab:fooling}), which is $72.62\%$ by VGG-19.  
This again emphasizes the effectiveness of the proposed framework and objectives to simultaneously model perturbations for classifiers with significant architectural differences.

\subsection{Black-box attacks for ensemble generator}
\label{subsec:black-box-ensemble}
In this subsection we present the black-box fooling performance of the learned generator for an ensemble of target CNNs. We have learned the ensemble generator $G_E$ with an ensemble of $4$ (VGG-F, GoogLeNet, VGG-$16$, and ResNet-$50$) target CNNs leaving CaffeNet, VGG-$19$, and ResNet-$152$. In Table~\ref{rebut-tab:black-box-ensemble} we report the mean black-box fooling rate (Mean BBFR) obtained by the learned perturbations computed over the three left out models. For comparison, we also present the Mean BBFR achieved by the generators learned for those individual target CNNs computed over the left out $3$ models. Owing to the ensemble of targets, generator $G_E$ learns more general perturbations compared to the individual generators and achieves higher fooling rate compared to the individual targets.
\vspace{-0.2cm}
\begin{table}[h]
\centering
\caption{Generalizability of the perturbations learned by the ensemble generator $(G_E)$.}
\label{rebut-tab:black-box-ensemble}
\begin{tabular}{|l|l|l|l|l|l|}
\hline
\multicolumn{1}{|l|}{}      &  $G_{VF}$ &$G_G$& $G_{V16}$ & $G_{R50}$ & $G_E$ \\ \hline
Mean BBFR & 60.63  &60.15    & 71.26  & 61.87    &   \textbf{76.40} \\ \hline
\end{tabular}
\end{table}
\section{Related Works}
\label{sec:relworks}
\textbf{Adversarial perturbations}~\cite{intriguing-iclr-2014} have been a tantalizing revelation about machine learning systems. Specifically, the deep neural network based learning systems~(\eg, \cite{explainingharnessing-iclr-2015,deepfool-cvpr-2016,distillation-sp-2016}) are also shown to be vulnerable to these structured perturbations. Their ability to generalize to unknown models enables simple ways~(\eg, \cite{explainingharnessing-iclr-2015}) to launch black-box attacks that fool the deployed systems. Further, existence of image-agnostic perturbations~\cite{universal-cvpr-2017,mopuri-bmvc-2017,gduap-arxiv-2018}
along with their cross model generalizability exposes the weakness of the current day deep learning models. Differing from previous works~\cite{universal-cvpr-2017,mopuri-bmvc-2017}, our work proposes a novel, yet simple and effective objective that enables to learn image-agnostic perturbations. Although the existing objectives successfully craft these perturbations, they do not attempt to capture the space of such perturbations. Unlike the existing works, the proposed method learns a generative model that can capture the space of the image-agnostic perturbations for a given classifier. To the best of our knowledge, the only work which aims to learn a neural network for generating adversarial perturbations via simple feed-forwarding is presented by Baluja~\textit{et al.}~\cite{atn-aaai-2018}. They present a neural network which transforms an image into its corresponding adversarial sample. Note that it generates image specific perturbations and doesn't aim to model the distribution like us.
\\
\textbf{Generative Adversarial Network (GAN):} 
Goodfellow~\textit{et al.}~\cite{gan-nips-2014} and Radford~\textit{et al.}~\cite{dcgan-arxiv-2015} have shown that GANs can be trained to learn a data distribution and to generate samples from it. Further image-to-image conditional GANs have led to improved generation quality~\cite{isola-cvpr2017,zhu-iccv-2017}. Inspired from GAN framework, we have proposed a neural network architecture to model the distribution of universal adversarial perturbations for a given classifier. The discriminator $(D)$ part of the typical GAN is replaced with the trained classifier to be fooled $(f)$. Only the generator $(G)$ part is learned to generate perturbations to fool the discriminator. Also, as we don't have to train $D$, samples of the target data (i.e., perturbations) are not presented during the training. Through a pair of effective loss functions $L_f \text{ and } L_d$, the proposed framework models the perturbations that fool a given classifier. 
\section{Conclusion}
\label{sec:conclu}
\vspace{-0.2cm}
In this paper, we have presented the first ever generative approach to model the distribution of adversarial perturbations for a given CNN classifier. We propose a GAN inspired framework, wherein we successfully train a generator network that captures the unknown target distribution without any training samples from it. The proposed objectives naturally exploit the attributes of the samples (\eg, to be able to fool the target CNN) in order to model their distribution. However, unlike the typical GAN training that deals with a pair of conflicting objectives, our approach has a single well behaved optimization (only $G$ is trained).

The ability of our method to generate perturbations with state-of-the-art fooling rates and surprising cross-model generalizability highlights severe susceptibilities of the current deep learning models. However, the proposed framework to model the distribution of perturbations also enables to conduct formal studies towards building robust systems. For example, Goodfellow \textit{et al.}~\cite{explainingharnessing-iclr-2015} introduced adversarial training as a means to learn robust models and Tramer \textit{et al.}~\cite{ensembleAT-iclr-2018} extended it to ensemble adversarial training, which require a large number of adversarial samples. In addition, the defence becomes more robust if those samples exhibit diversity and allow the model to fully explore the space of adversarial examples. While the existing methods are limited by both generation speed and instance diversity, our method, after modelling, almost instantly produces adversarial perturbations with lots of variety. We have also shown that our approach can efficiently model the perturbations that simultaneously fool multiple deep models.
\onecolumn
\begin{center}
\textbf{\large Appendix }
\end{center}
\setcounter{section}{0}
\setcounter{figure}{0}
\setcounter{equation}{0}
\section{Traversing the manifold of perturbations}
\label{sec:interpolation}
\begin{figure*}[ht]
\centering
\noindent\begin{minipage}{\textwidth}
  \centering
  \begin{minipage}{.18\textwidth}
  	\centering
    \includegraphics[width=\linewidth]{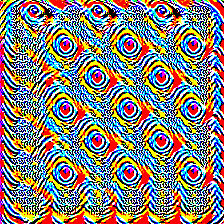}\
    $0.0*z_1+1.0*z_2:89.87$
  \end{minipage}
   \begin{minipage}{.18\textwidth}
   	\centering
    \includegraphics[width=\linewidth]{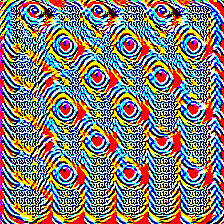}\
    $0.1*z_1+0.9*z_2:89.76$
  \end{minipage}
  \begin{minipage}{.18\textwidth}
  	\centering
    \includegraphics[width=\linewidth]{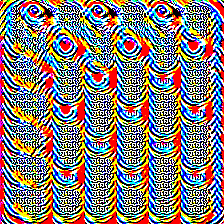}\
    $0.2*z_1+0.8*z_2:90.58$
  \end{minipage}
   \begin{minipage}{.18\textwidth}
   	\centering
    \includegraphics[width=\linewidth]{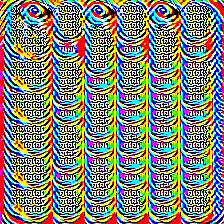}\
    $0.3*z_1+0.7*z_2:91.28$
  \end{minipage}
  \begin{minipage}{.18\textwidth}
  	\centering
    \includegraphics[width=\linewidth]{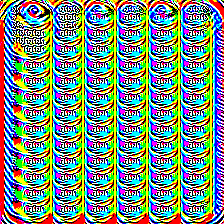}\
     $0.4*z_1+0.6*z_2:91.96$
  \end{minipage}
  \vspace{0.002\textwidth}
\end{minipage}
\noindent\begin{minipage}{\textwidth}
  \centering
  \begin{minipage}{.18\textwidth}
  	\centering
    \includegraphics[width=\linewidth]{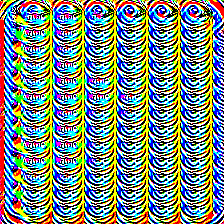}\
    $0.5*z_1+0.5*z_2:89.28$
  \end{minipage}
   \begin{minipage}{.18\textwidth}
   	\centering
    \includegraphics[width=\linewidth]{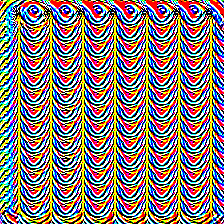}\
    $0.6*z_1+0.4*z_2:91.06$
  \end{minipage}
  \begin{minipage}{.18\textwidth}
  	\centering
    \includegraphics[width=\linewidth]{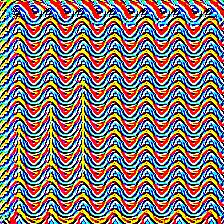}\
    $0.7*z_1+0.3*z_2:91.12$
  \end{minipage}
   \begin{minipage}{.18\textwidth}
   	\centering
    \includegraphics[width=\linewidth]{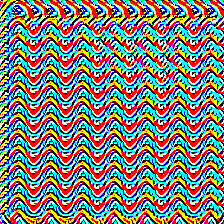}\
    $0.8*z_1+0.2*z_2:91.28$
  \end{minipage}
  \begin{minipage}{.18\textwidth}
  	\centering
    \includegraphics[width=\linewidth]{figures/resnet152_interpolation/interpolation_resnet152_10.png}\
     $0.9*z1 + 0.1*z_2:90.04$
  \end{minipage}
  \vspace{0.002\textwidth}
\end{minipage}
\vspace{0.1cm}
\caption{Interpolation between a pair of points in $Z$ space show that the distribution learned by our generator has smooth transitions. The figure shows the perturbations corresponding to $10$ points on the line joining a pair of random points $(z_1 \text{ and } z_2)$ in the latent space. Note that these perturbations are learned to fool the GoogLeNet~\cite{googlenet-cvpr-2015} architecture. Below each perturbation, the corresponding fooling rate obtained over $50000$ images from ILSVRC $2014$ validation images is mentioned. This shows the fooling capability of these intermediate perturbations is also high and it is similar at different locations in the learned distribution of perturbations.}
\label{fig:interpolation-appendix}
\end{figure*}

In this section, we provide additional results for traversing the manifold of adversarial perturbations. On the line joining two random points $z_1$ and $z_2$ in the latent space, we consider $10$ intermediate points. We transform these points into corresponding perturbations by forwarding through the learned $G$ for GoogLeNet~\cite{googlenet-cvpr-2015}. Fig.~\ref{fig:interpolation-appendix} shows those $10$ generated perturbations. Similar to the results presented on ResNet-152~\cite{resnet-cvpr-2016} classifier shown in the main draft (refer to sec.3.3), the perturbations change smoothly between any pair of consecutive points. The circular patterns in the first image of the top row slowly disappear by the $4^{th}$ image (top row) while curtain like patterns emerge from $3^{rd}$ image of top row, which are clearly visible by the end of top row. Proceeding further on the path generates wave like patterns shown in middle of bottom row which transform to a new pattern.  Also, the fooling rates are high and consistent along the path. This process again emphasizes that the proposed model learns relevant representations.

\section{Generator details}
\label{sec:g-details}
In this section, we explain the architecture of the generator network in the proposed framework that learns the manifold of adversarial perturbations for a given target classifier. Table~\ref{tab:gen-arch} presents the details of the generator network which basically includes an initial fully connected layer followed by several deconvolutional layers to map the latent space vectors to image size perturbations. The architecture is adapted from~\cite{improvedgans-nips-2016} which presents improved techniques to GAN training such as virtual batch normalization to better stabilize the training of the generator network. Note that for all our experiments, the generator architecture is fixed independent of the target CNN under attack. 
\begin{table}[]
\centering
\caption{Overview of the layers in the generator network to model the adversarial perturbations by transforming a latent space vector $z$ of dimension $10$ to a perturbation of size $224 \times 224$.}
\label{tab:gen-arch}
\begin{tabular}{|c|}
\hline
\multicolumn{1}{|c|}{\textbf{Generator}}                \\ \hline
FC (10, 64*7*4*4)                  \\ \hline
Virtual BN, ReLU \\ \hline
Deconv (1,7,7,64*4)               \\ \hline
Virtual BN, ReLU \\ \hline
Deconv (1,14,14,64*2)             \\ \hline
Virtual BN, ReLU \\ \hline
Deconv (1,28,28,64*1)             \\ \hline
Virtual BN, ReLU \\ \hline
Deconv (1,56,56,64*1)             \\ \hline
Virtual BN, ReLU \\ \hline
Deconv (1,112,112,64*1)           \\ \hline
Virtual BN, ReLU \\ \hline
Deconv (1,224,224,64*1)           \\ \hline
Virtual BN, ReLU \\ \hline
Deconv (1,224,224,3)              \\ \hline
\textit{10*tanh}                  \\ \hline
\end{tabular}
\end{table}

\section{Modelling adversaries for Multiple target classifiers}
\label{sec:multipletarget}
\begin{figure*}[h]
\centering
\noindent\begin{minipage}{\textwidth}
  \centering
  \begin{minipage}{.24\textwidth}
  	\centering
    \includegraphics[width=\linewidth]{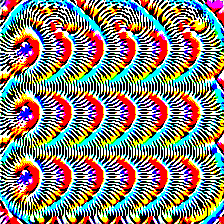}\
    
  \end{minipage}
   \begin{minipage}{.24\textwidth}
   	\centering
    \includegraphics[width=\linewidth]{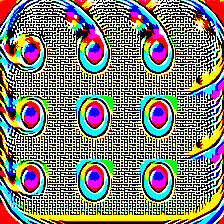}\
    
  \end{minipage}
  \begin{minipage}{.24\textwidth}
  	\centering
    \includegraphics[width=\linewidth]{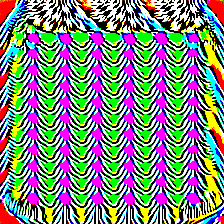}\
    
  \end{minipage}
   \begin{minipage}{.24\textwidth}
   	\centering
    \includegraphics[width=\linewidth]{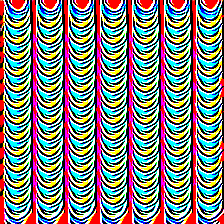}\
    
  \end{minipage}

  \vspace{0.002\textwidth}
\end{minipage}
\vspace{0.1cm}

\caption{Sample perturbations obtained from the distribution of perturbations learned for multiple target classifiers. The $G$ is learned to craft perturbations which can simultaneously fool the $7$ CNNs mentioned in the main draft. Note that the perturbations are of wide variety.}
\label{fig:multi-target-pb}
\end{figure*}

In this section, we present sample perturbations obtained for multi-target case, i.e., the generator is learned to model the adversaries that can simultaneously fool multiple target classifiers. Fig.~\ref{fig:multi-target-pb} shows sample perturbations captured by the learned distribution. Note that the perturbations exhibit significant visual diversity. Also, in section~\ref{sec:multi-target-diversity} of this appendix, we show that the perturbations exhibit diversity in terms of the predicted labels after adding to clean images.

\section{Representations for diversity objective}
\label{subsec:ablation-representations}
The proposed loss 
has a diversity component in order for the generative model $G$ to be able to capture all the variations in the distribution of perturbations. The diversity objective $L_d$ 
indirectly increases the distance between any pair of perturbations generated in a mini-batch $(\delta_i \text{ and }{\delta}'_i)$. The objective projects the feature representations belonging to the \textit{adversarial} and \textit{shuffled adversarial} samples apart. 
As the transformations learned by layers of the target CNN $(D)$ are fixed, the objective helps $G$ to generate a diverse set of perturbations that fool the target. 

In this subsection, we investigate the effectiveness of multiple feature representations to encourage the diversity in the modelling of the perturbations. We have worked with VGG-F~\cite{vggf-bmvc-2014} model for this analysis and considered conv4, conv5 and softmax layers. For separating features at conv4 and conv5 layers we have employed Euclidean distance and for features obtained at softmax layer, cosine distance is maximized. Table~\ref{tab:ablation-representations} shows the fooling rates obtained by the proposed objective. We have run the optimization $10$ times and report the mean fooling rates along with their standard deviations. Note that the fooling performance is almost same (less than $2$ \% difference) for all the three representations. Also, the standard deviations are very less indicating the effectiveness of the fooling objective and faithful modelling of the distribution. However, upon visual investigation, we observe that the variations captured via separating the softmax representations is relatively high. 
Thus, we chose to work with representations at softmax layer to learn the generative model of perturbations in all our experiments. 

\begin{table*}[h]\centering
\caption{Ablation results for different representations to encourage capturing the diversity in the distribution of perturbations for VGG-F~\cite{vggf-bmvc-2014} model. Note the numbers are the fooling rates obtained over a held-out set across $10$ runs of the optimization. Right most column reports the mean fooling rate along with the standard deviation.}
\label{tab:ablation-representations}
\begin{tabular}{|l|l|l|l|l|l|l|l|l|l|l|l|}
\hline
\backslashbox{Feature \\ + Metric}{\#Run}  & 1     & 2     & 3     & 4     & 5     & 6     & 7     & 8     & 9     & 10    & Mean (std.) \\ \hline
Conv4 + $L_2$             & 93.82 & 93.76 & 93.95 & 93.81 & 93.92 & 93.86 & 93.88 & 93.91 & 93.87 & 93.90 & 93.86 (0.055)  \\ \hline
Conv5 + $L_2$             & 94.77 & 94.75 & 94.86 & 94.76 & 94.82 & 94.83 & 94.95 & 94.70 & 94.73 & 94.87 & 94.80 (0.072)  \\ \hline
Softmax + cosine  & 93.32 & 93.63 & 93.62 & 93.60 & 93.52 & 93.74 & 93.43 & 93.64 & 93.66 & 93.57 & 93.57 (0.11)  \\ \hline
\end{tabular}
\end{table*}
\section{Sample benign and corresponding perturbed images}
\begin{figure*}[h]
\centering
\noindent\begin{minipage}{\textwidth}
  \centering
  \begin{minipage}{.18\textwidth}
  	\centering
    \includegraphics[width=\linewidth]{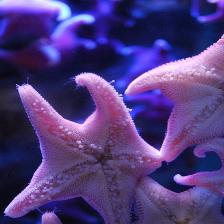}\
    $Star fish$
  \end{minipage}
   \begin{minipage}{.18\textwidth}
   	\centering
    \includegraphics[width=\linewidth]{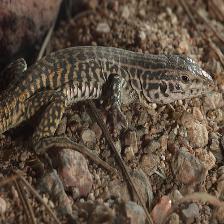}\
    $Whiptail$
  \end{minipage}
  \begin{minipage}{.18\textwidth}
  	\centering
    \includegraphics[width=\linewidth]{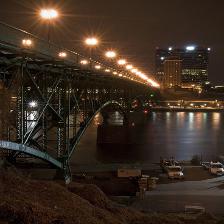}\
    $Steel arch bridge$
  \end{minipage}
   \begin{minipage}{.18\textwidth}
   	\centering
    \includegraphics[width=\linewidth]{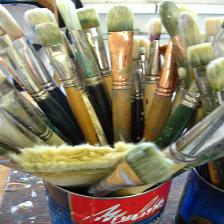}\
    $Paintbrush$
  \end{minipage}
  \begin{minipage}{.18\textwidth}
  	\centering
    \includegraphics[width=\linewidth]{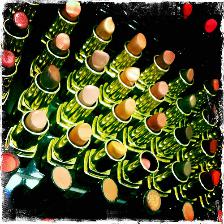}\
     $Lipstick$
  \end{minipage}
  \vspace{0.002\textwidth}
\end{minipage}
\noindent\begin{minipage}{\textwidth}
  \centering
  \begin{minipage}{.18\textwidth}
  	\centering
    \includegraphics[width=\linewidth]{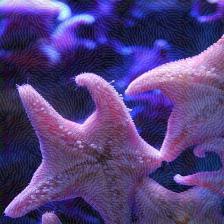}\
    $Gibbon$
  \end{minipage}
   \begin{minipage}{.18\textwidth}
   	\centering
    \includegraphics[width=\linewidth]{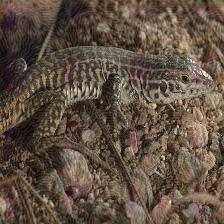}\
    $Sidewinder$
  \end{minipage}
  \begin{minipage}{.18\textwidth}
  	\centering
    \includegraphics[width=\linewidth]{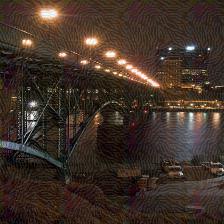}\
    $Porcupine$
  \end{minipage}
   \begin{minipage}{.18\textwidth}
   	\centering
    \includegraphics[width=\linewidth]{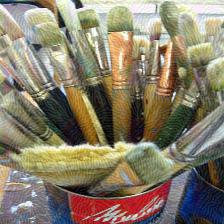}\
    $Ear$
  \end{minipage}
  \begin{minipage}{.18\textwidth}
  	\centering
    \includegraphics[width=\linewidth]{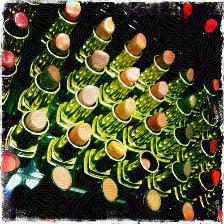}\
     $Tray$
  \end{minipage}
  \vspace{0.002\textwidth}
\end{minipage}
\vspace{0.1cm}
\caption{Set of benign and corresponding perturbed images for CaffeNet~\cite{caffe-acmmm-2014}. Top row shows the benign samples with their ground truth labels below. Bottom row shows the predicted label on the corresponding perturbed images.}
\label{fig:perturbations-single-net}
\end{figure*}

 Fig.~\ref{fig:perturbations-single-net} shows a set of benign (top row) and corresponding adversarial images (bottom row) for CaffeNet~\cite{caffe-acmmm-2014}. Labels provided in the top row are ground truth and in the bottom row are the predicted ones by CaffeNet. Note that different perturbations are added to each of the images which result in different predicted labels. 

\section{Weighting the individual loss components}
\label{sec:weight-loss-comps}
In this section we perform experiments to understand how weighting of the two loss components i.e., diversity objective $(L_d)$ and fooling objective $(L_f)$ affects the learning. The target architecture we consider for this set of ablations is ResNet-50 architecture trained on ILSVRC-$2014$ data. We vary the value of $\lambda$ (weight given to the diversity objective) from $0.1$ to $10$ in steps of $10$ and separately learn three generators. Table~\ref{tab:weight-loss-comps} presents the corresponding fooling rate and diversity measure for the learned generators. Note that the diversity measure is the number of labels that account for $95\%$ of the predicted labels after adding the perturbations (Refer sec~$3.2.2$ of the main draft). Also, these numbers are computed on the $50000$ validation images of ILSVRC-$2014$ data. Results demonstrate that both the measures clearly reflect the relative weights given to the corresponding loss components. That is, fooling rate increases from $76.26\%$ to $89.04\%$ as the weight given to $L_f$ increases, at the same time the number of dominant labels decreases from $348$ to $253$. This observation clearly demonstrates that the weights given to the individual loss components correlate with the resulting performance. We can treat the weight $\lambda$ as an adjustable hyper parameter in the system that can control the generated perturbations from \textit{high fooling - limited variety} to \textit{less fooling - wide variety}. 

\begin{table}
\centering
\caption{Effect of weighting loss components while learning the generator. Note that weights given to the components correlate to the resulting performance.}
\label{tab:weight-loss-comps}
\begin{tabular}{ccc}\\ \hline
Loss (L)  & Fooling rate & Diversity measure \\ \hline
$ L_f + 10 \times L_d$ & 76.26        & 348       \\
$L_f + L_d$     & 86.64 &  315     \\   
$ L_f + 0.1 \times L_d$ & 89.04  & 253  \\ \hline 
\end{tabular}
\end{table}

\section{Diversity in perturbations learned by a generator}
\label{sec:multi-target-diversity}
\begin{figure*}[h]
\centering
  \centering
  \begin{minipage}{.18\textwidth}
  	\centering
    \includegraphics[width=\linewidth]{figures/perturbed_samples/test2.jpg}\
    {{\color{green}Whiptail}}
  \end{minipage}
   \begin{minipage}{.18\textwidth}
   	\centering
    \includegraphics[width=\linewidth]{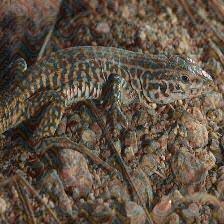}\
    {\color{red}Alligator lizard}
  \end{minipage}
  \begin{minipage}{.18\textwidth}
  	\centering
    \includegraphics[width=\linewidth]{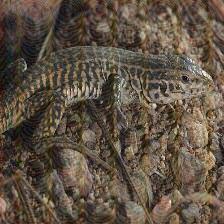}\
    {\color{red}Sidewinder}
  \end{minipage}
   \begin{minipage}{.18\textwidth}
   	\centering
    \includegraphics[width=\linewidth]{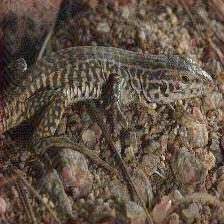}\
    {\color{red}Banded gecko}
  \end{minipage}
\begin{minipage}{.18\textwidth}
   	\centering
    \includegraphics[width=\linewidth]{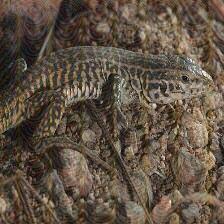}\
    {\color{red}Common newt}
  \end{minipage}
\vspace{0.1cm}

\caption{Sample benign image (extreme left) and four perturbed images obtained by adding different perturbations learned the proposed generator $(G)$. Note that the $G$ is learned to fool all the $7$ target CNNs and predictions are obtained by GoogLeNet. The ground truth for the benign image is mentioned in green below it. Predicted labels mentioned below the corresponding images in red and all of them are different.}
\label{fig:multi-target-diversity}
\end{figure*}

In this section we present example perturbed images crafted by adding different perturbations learned by a Generator $(G)$ to a given benign image. Note that we have learned the $G$ to model the perturbations that simultaneously fool all the $7$ target CNNs mentioned in the main draft (refer to sec.3.4 in the main draft). The predictions are obtained by GoogLeNet~\cite{googlenet-cvpr-2015}. Fig.~\ref{fig:multi-target-diversity} shows the benign and corresponding perturbed images crafted by adding $4$ random perturbations generated by the $G$. Leftmost image is a benign sample and the following four images are perturbed samples. The predicted labels are mentioned below the corresponding images. Note that all the four predictions are different and this demonstrates that the perturbations $(\delta)$ learned by a generator $(G)$ exhibit diversity in terms of the predicted labels. Importantly, we observe similar diversity for all the generators trained on different target CNNs.

\twocolumn
\bibliographystyle{ieee}
\bibliography{mybibliography}

\end{document}